# Enhancing Autonomous Vehicle Perception in Adverse Weather through Image Augmentation during Semantic Segmentation Training

Ethan Kou[1,2]    Noah Curran[3#]

[1] Henry M. Gunn High School, Palo Alto, California
[2] Electrical Engineering and Computer Sciences, UC Berkeley
[3] Computer Science and Engineering, University of Michigan, [#]Advisor

https://github.com/BubblyBingBong/AugmentationSegmentation

**Abstract.** Robust perception is crucial in autonomous vehicle navigation and localization. Visual processing tasks, like semantic segmentation, should work in varying weather conditions and during different times of day. Semantic segmentation is where each pixel is assigned a class, which is useful for locating overall features (1). Training a segmentation model requires large amounts of data, and the labeling process for segmentation data is especially tedious. Additionally, many large datasets include only images taken in clear weather. This is a problem because training a model exclusively on clear weather data hinders performance in adverse weather conditions like fog or rain. We hypothesize that given a dataset of only clear days images, applying image augmentation (such as random rain, fog, and brightness) during training allows for domain adaptation to diverse weather conditions. We used CARLA, a 3D realistic autonomous vehicle simulator, to collect 1200 images in clear weather composed of 29 classes from 10 different towns (2). We also collected 1200 images of random weather effects. We trained encoder-decoder UNet models to perform semantic segmentation. Applying augmentations significantly improved segmentation under weathered night conditions ($p < 0.001$). However, models trained on weather data have significantly lower losses than those trained on augmented data in all conditions except for clear days. This shows there is room for improvement in the domain adaptation approach. Future work should test more types of augmentations and also use real-life images instead of CARLA. Ideally, the augmented model meets or exceeds the performance of the weather model.

## 1. INTRODUCTION

Detection and perception are crucial tasks for autonomous vehicles because they allow for accurate driving in a complex environment. Semantic segmentation is one example of a perception task. In semantic segmentation, each pixel in the image is given a class label (1). For example, cars are all given an ID, and roads are given a separate ID (Figure 1). Using this segmented image, autonomous vehicles can locate important landmarks to perform autonomous actions.

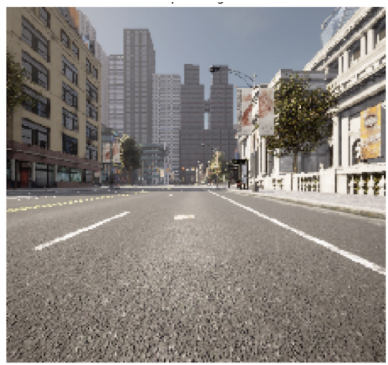 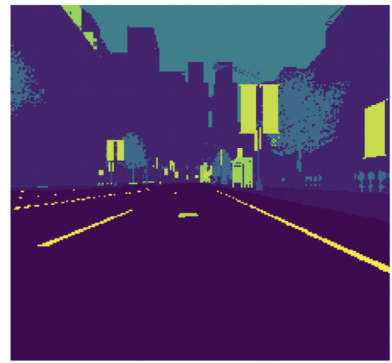

Figure 1: Semantic segmentation example. RGB image on the left and semantic segmentation mask on the right, taken from our dataset.

To train a robust segmentation model, large amounts of images and labels need to be collected. Labels are especially tedious to create for image segmentation because each pixel's class needs to be hand-labeled. Furthermore, most large datasets, such as CityScapes, are composed of images collected in clear weather conditions (3). This results in lower segmentation accuracy when the model is tested on images in adverse weather conditions. For example, rain can obstruct parts of the image, and fog can cause blurring.

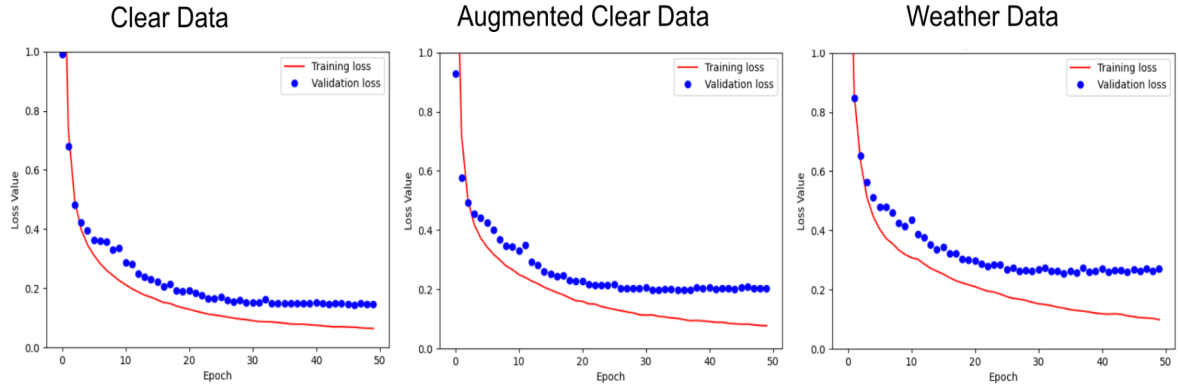

Figure 2: Training vs validation loss for M1, M2, and M3. M1 was trained for 50 epochs on clear-day images (D1). M2 was trained for 50 epochs on augmented clear-day images (D1). M3 was trained for 50 epochs on weather images.

There have been many approaches to the challenge of performing detection in various weather conditions. DFR-TSD improved traffic sign detection by training models to reverse weather effects on images (4). DGSS generated adversarial styles to improve nighttime segmentation (5). Some papers have also introduced tests for the robustness of models using various image transformations (6). Few papers discuss autonomous vehicle semantic segmentation domain adaptation across both time and weather when only given clear weather data. Our hypothesis is that augmenting clear weather data in training improves semantic segmentation in adverse weather conditions. We believe the augmented data can train the model to extract features invariant to the domain.

## 2. RESULTS

We trained three semantic segmentation models with the UNet architecture. The first model, M1, was trained on clear weather images. M2 was trained on augmented clear weather images, and M3 was trained on weather images. The training and validation loss of the three models are shown in (Figure 2). The loss value measures how much the prediction differs from the actual, so a lower loss is better. We used the TensorFlow SparseCategorialCrossentropy loss. The performance of each model in varying weathers and times of day is in Tables 1 and 2.

|    | DC     | DR     | DW     | NC     | NR     | NW     | W      |
|----|--------|--------|--------|--------|--------|--------|--------|
| M1 | 0.1423 | 0.525  | 0.2171 | 1.3228 | 1.4114 | 1.0673 | 0.5045 |
| M2 | 0.1535 | 0.4368 | 0.2252 | 0.991  | 1.2082 | 0.846  | 0.4369 |
| M3 | 0.1919 | 0.2588 | 0.195  | 0.4341 | 0.5546 | 0.4455 | 0.2846 |

Table 1: Loss of models in various test sets. The dataset name follows the convention of {D/N}{C/R/W}, where D = day, N = night, C = clear, R = rainy, and W = random weather and time.

With augmentation, the accuracy for random weather conditions improved from 0.8600 to 0.8822, and the loss decreased from 0.5045 to 0.4369 (Table 1, 2). This indicates that image augmentation may generally improve model performance. To show more concrete statistical evidence, we conducted a 2-sample t-test which showed that models trained on augmented data had significantly lower loss than models trained on clear data when tested on rainy night data and random weather night data. ($p < 0.001$, 2-sample t-test, Table 3).

|    | DC     | DR     | DW     | NC     | NR     | NW     | W      |
|----|--------|--------|--------|--------|--------|--------|--------|
| M1 | 0.9588 | 0.8606 | 0.9381 | 0.6332 | 0.6338 | 0.7145 | 0.8600 |
| M2 | 0.955  | 0.8831 | 0.937  | 0.715  | 0.7046 | 0.7754 | 0.8822 |
| M3 | 0.9455 | 0.9279 | 0.9435 | 0.8844 | 0.8601 | 0.8829 | 0.9231 |

Table 2: Accuracy of models in various test sets. The dataset name follows the convention of {D/N}{C/R/W}, where D = day, N = night, C = clear, R = rainy, and W = random weather and time.

|  | DC | DR | DW | NC | NR | NW | W |
|---|---|---|---|---|---|---|---|
| Clear Mean | 0.1878 | 0.5270 | 0.2522 | 0.9467 | 1.2077 | 0.8832 | 0.4600 |
| Clear Stdev | 0.0064 | 0.0811 | 0.0131 | 0.0877 | 0.2240 | 0.1002 | 0.0403 |
| Augmented Mean | 0.2337 | 0.4752 | 0.3052 | 0.9393 | 0.8710 | 0.7312 | 0.4299 |
| Augmented Stdev | 0.0085 | 0.0863 | 0.0098 | 0.1131 | 0.1035 | 0.0530 | 0.0228 |
| Weather Mean | 0.2319 | 0.2882 | 0.2396 | 0.4222 | 0.5112 | 0.4318 | 0.3008 |
| Weather Stdev | 0.0076 | 0.0080 | 0.0068 | 0.0119 | 0.0149 | 0.0165 | 0.0078 |
| Augmented lower loss than clear, p-value | 1.00000 | 0.09175 | 1.00000 | 0.43614 | 0.00021 | 0.00025 | 0.02742 |
| Weather lower loss than augmented, p-value | 0.31303 | <0.0001 | <0.0001 | <0.0001 | <0.0001 | <0.0001 | <0.0001 |

Table 3: Loss and single-tailed 2-sample t-test p-values of 10-fold models in various test sets. The table stores the mean and standard deviation of the loss of the 10-fold models for the clear, augmented, and weather datasets. The p-value for the augmented dataset models having lower loss than the clear dataset models and the p-value for the weather dataset models having lower loss than the augmented dataset models are written. The dataset name follows the convention of {D/N}{C/R/W}, where D = day, N = night, C = clear, R = rainy, and W = random weather and time.

The randomness and diversity of augmentations could explain this trend, allowing the model to adapt better to adverse lighting conditions during the night. There is also less "noise" in the augmented model prediction than in the clear weather prediction (Figure 3). The loss of the weather data models is significantly lower than that of the augmented data models on all 7 test sets except DC ($p < 0.001$, 2-sample t-test, Table 3). Therefore, there is room for improvement in the domain adaptation approach. The optimal domain adaptation approach would meet or exceed the performance of a model trained on weather data when only given ground truth clear data.

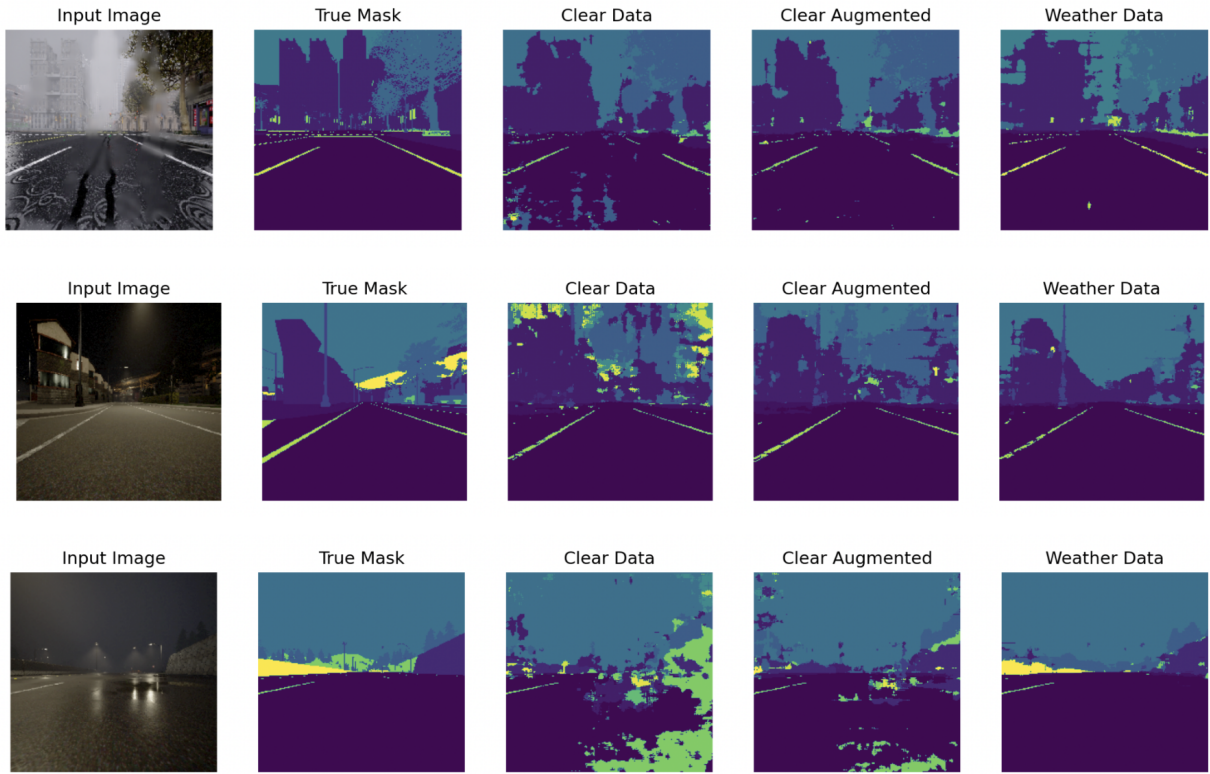

Figure 3: Example predictions on 3 images. M1, M2, and M3 were tested on various images in our dataset.

## 3. DISCUSSION

The results show evidence that when only training on clear-day images, applying image augmentation improves model performance during adverse weather conditions. There is also statistically significant evidence that augmentations decrease test loss in night-time rain or night-time random weather conditions.

Despite the positive results, there are many limitations and possible improvements. First, custom augmentations outside of the Albumentation library should also be used. Albumentations is a library we used that has image augmentation capabilities (7). Currently, augmentations we use include random rain, random fog, and some others. We review each augmentation in more detail in the Materials and Methods section. Although these augmentations apply weather variation to the image, they are weather-specific augmentations. Weather-specific models are less effective because the model might have a weaker result when tested in unseen weather like snow. As future work, the augmentations should be non-weather-specific and still introduce realistic weather variation. For example, it's possible that a combination of blur, noise, and other image filters could span the possible image transformations caused by weather. This could improve the detection in a wider range of weathers instead of the specific weathers that the augmentations tailor.

The significance testing method could also be improved. Since we used 10-fold cross-validation for training, some models will share data, making the loss of the models not completely independent from each other. This can cause problems with the accuracy of the significance test because the t-test assumes that all samples are independent. Future work would ensure every sample is independent. This can be done by acquiring a dataset large enough so that the 10 models that are trained have no shared data.

A simulation limitation is that CARLA, the autonomous vehicle simulation we used, does not include weather options such as snow (2). Future work can be to use a dataset of more diverse weather conditions. This can be done in simulation, but simulations still have drawbacks. For our data collection in CARLA, the car always spawns perfectly aligned on the road, and many factors of random noise are reduced. This level of perfection is absent in real life, so extrapolating our approach to real-life datasets is an important next step. This paper serves as a stepping point for the end goal of training a model to effectively perform segmentation in real-life adverse weather when only given data in clear day weather.

## 4. MATERIALS AND METHODS

### Data Collection

We collected data using CARLA, a widely used 3D autonomous vehicle simulator. CARLA has numerous towns, each with its own architecture and location. Some towns are urban cities with skyscrapers, while others are rural farmlands with barns and crops.

CARLA sensors can be used to capture a scene in RGB and semantic segmentation. We attach both cameras to the front of the vehicle, so all images are captured on the road. The semantic segmentation is split into 29 different classes. View CARLA documentation for details on each class. We collected 2 training datasets of 1200 images each. The D1 dataset includes only images taken in clear weather using the weather ClearNoon preset. The D2 dataset includes randomly generated times of day and weather from the CARLA presets.

The procedure for generating each dataset is first setting the world to Town 1 and teleporting the vehicle to a random location on the road. Then, the weather will be adjusted according to the desired dataset, and the current snapshot will be recorded from the RGB and semantic segmentation camera. Repeat this until 150 data points are recorded for Town 1. Then, repeat everything for Towns 2-7 and 10. In total, each dataset has 1200 images because 150 data points are recorded for each of the 8 towns.

We generated seven test sets of 400 images, each under various weather conditions. We followed the same data collection procedure as the training set but only collected 50 images per town, for a total of 400 images per training set. The seven datasets are collected on clear days (DC), rainy days (DR), random weather days (DW), clear nights (NC), rainy nights (NR), random weather nights (NW), and random everything (W).

### Image Augmentation

The goal of image augmentations is to teach the model to extract features invariant to the domain. To do this, we applied a series of weather augmentations to each image. We used the Albumentations library to make applying various types of image augmentations easier. We used the Albumentations library to generate random rain, random sun flare, random fog, random RGB shift, and random gamma. Each of the 5 augmentations has a 0.5 probability of being applied to an image. When these augmentations are compounded, they produce a diverse range of images that are then inputted into the model for training (Figure 4).

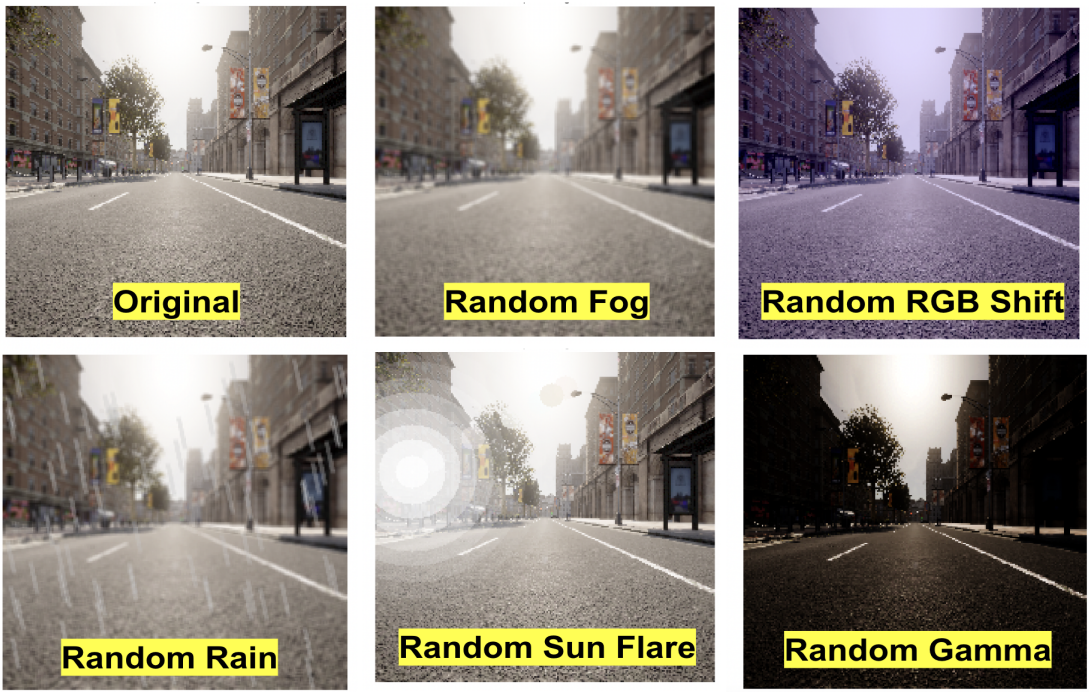

Figure 4: Augmentation Examples. Fog adds random blurring. RGB shift randomly shifts all pixel values. Random rain adds rain streaks. Random sun flare adds bright spots. Random gamma adjusts overall image brightness.

### *Overall Model Training and Evaluation*

We first trained 3 models. The first model is trained on D1, which has all clear weather images. The second model is trained on D1 after applying image augmentations. The third model is trained on D2, which has images in random weather and during different times of day.

We utilized a UNet architecture with a MobileNetV2 backbone pre-trained on ImageNet. The UNet is an encoder-decoder convolutional neural network that can be used for semantic segmentation. The encoder (contracting path) detects all complexities of image features using convolutional and max pooling layers, and the decoder (expansive path) upscales these detected features to segmentation masks using convolutional and unpooling layers (8). There are skip connections between layers of the encoder and decoder to allow for reliable segmentation across both high-level features, like the location of an edge, and low-level features, like the class of an object. The input images are resized to 224x224. For all models, we trained for 50 epochs. We randomly selected 1000 train samples and 200 validation samples.

### *Significance Testing*

To determine if image augmentations during training create a statistically significant decrease in the loss value, we trained 10 models each on D1, augmented D1, and D2 for 30 models total. We used 10-fold cross-validation for each dataset. For example, the first model is trained on all 1200 images except the first 120. The second model is trained on all 1200 images except the second 120, and so on. Each model was trained for 20 epochs because that number reduces total training time and is still high enough to prevent underfitting. We evaluated the loss of each of the 20 models on all 7 test sets.

For every test set, we have 10 loss values for each of the 3 datasets. To determine the p-value, we run a single-tailed 2-sample t-test on corresponding sets of 10 loss values for each test set.

# ACKNOWLEDGMENTS

Thanks to my mentor Noah Curran for the guidance.

# APPENDIX

You can find the source code for this project at https://github.com/BubblyBingBong/AugmentationSegmentation